\documentclass[10pt,twocolumn,letterpaper]{article}

\usepackage{iccp}
\usepackage{times}
\usepackage{amsmath}
\usepackage{amssymb}
\usepackage{url}
\usepackage{graphicx}
\usepackage{caption}
\usepackage{fix-cm}
\usepackage{placeins}
\usepackage[toc,page]{appendix}
\usepackage{xcolor}
\usepackage{diagbox}

\iccpfinalcopy % *** Uncomment this line for the final submission

\def\iccpPaperID{37} % *** Enter the ICCP Paper ID here

\ificcpfinal\fi
\begin{document}

%%%%%%%%% TITLE
\title{Fast and Accurate Reconstruction of Compressed Color Light Field}

\author{Ofir Nabati, David Mendlovic and Raja Giryes \\
Tel Aviv University, Tel Aviv 6997801\\
{\tt\small \url{ofirnabati@mail.tau.ac.il}, \url{mend@eng.tau.ac.il}, \url{raja@tauex.tau.ac.il} }}

\twocolumn[{
\centering
\maketitle
\thispagestyle{empty}

%\begin{figure*}[h]
\begin{center}
\includegraphics[width=1\linewidth]{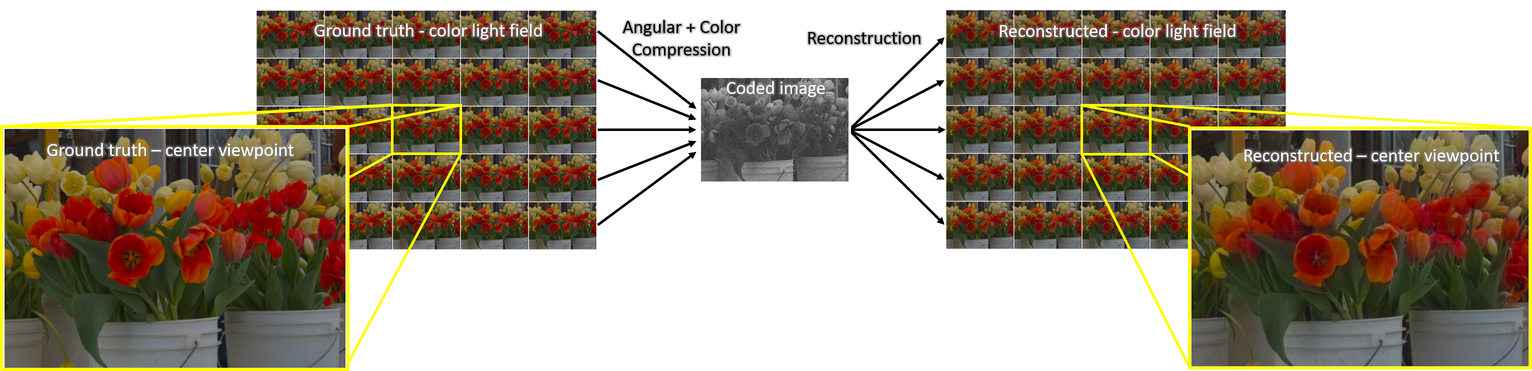}
  \captionof{figure}{Reconstruction of a color light field image from a 2D coded image projected at the sensor in a single shot. The compression of the color and angular information is done by the optical system, using a random coded color mask placed between the aperture and the sensor. We recover the full-color light field using a computationally efficient neural network.}
\label{fig:long}
\label{fig:onecol}
\end{center}
}]

%%%%%%%%% ABSTRACT
\begin{abstract}
Light field photography has been studied thoroughly in recent years. One of its drawbacks is the need for multi-lens in the imaging. To compensate that, compressed light field photography has been proposed to tackle the trade-offs between the spatial and angular resolutions. It obtains by only one lens, a compressed version of the regular multi-lens system. The acquisition system consists of a dedicated hardware followed by a decompression algorithm, which usually suffers from high computational time. In this work, we propose a computationally efficient neural network that recovers a high-quality color light field from a single coded image. Unlike previous works, we compress the color channels as well, removing the need for a CFA in the imaging system.
Our approach outperforms existing solutions in terms of recovery quality and computational complexity. We propose also a neural network for depth map extraction based on the decompressed light field, which is trained in an unsupervised manner without the ground truth depth map. 
\end{abstract}

%%%%%%%%% BODY TEXT
\section{Introduction}

While conventional 2D images contain only the RGB content of a given scene, light field images hold also the angular information. This allows performing challenging tasks such as refocusing and depth extraction, which are harder to be done using only the spatial information. One way of representing light field information is as a collection of 2D images, taken from multi-viewpoints \cite{levoy1996light}.

Light field images may be captured by various methods such as coded masks, coded apertures, microlenses and pinhole arrays. Due to limited sensor size, these systems suffer from a tradeoff between the spatial and angular resolution that usually results in a sparse number of viewpoints. To address this drawback, bulky imaging systems or array of sensors have been proposed \cite{wilburn2005high}. These solutions are either impractical or expensive and bulky as they require a large amount of storage and have a bigger size. 

Recently, Marwah {\em et al.} have introduced the concept of \textbf{compressive light field photography} \cite{marwah2013compressive}. 
Following the compressed sensing theory \cite{candes2008introduction,donoho2006compressed},
they reconstruct a high-resolution light field from its 2D coded projection measured by a single sensor. This theory guarantees under some conditions the recovery of a signal, which has a sparse representation in a given dictionary, from a relatively small number of linear measurements. The authors in \cite{marwah2013compressive} use a learned dictionary decomposed of light field atoms. Following this work, several improvements have been proposed for both the sensing system and the recovery algorithms \cite{chen2017light,gupta2017compressive,hirsch2014switchable}. Nevertheless, these methods still do not lead to real-time light field acquisition. Also, to the best of our knowledge, there is no reference to \textbf{color} compression in these works.

Based on the strategy of Marwah et al., we propose a novel system for reconstructing a full color light field from the compressed measurements acquired by a conventional camera with a random color mask. The light field estimation is performed by deep learning, which is an emerging machine learning methodology that solves a given problem by training a neural network using a set of training examples. It achieves state-of-the-art results in various computer vision and image processing tasks such as image classification \cite{krizhevsky2012imagenet,szegedy2015going}, semantic segmentation \cite{long2015fully}, image denoising \cite{remez2017deep}, image super-resolution \cite{ledig2016photo}, depth estimation \cite{flynn2016deepstereo}, etc. Neural networks have been also implemented to synthesize new light field views from few given perspectives \cite{kalantari2016learning} or even a single view \cite{srinivasan2017learning}. They have been used also to solve inverse problems with sparsity priors \cite{Giryes18Tradeoffs, gregor2010learning,sprechmann2015learning} and specifically compressed sensing problems  \cite{8122281,kulkarni2016reconnet}. These methods have shown to produce better and faster results than "classic" sparse coding techniques. 

In this work, we use deep learning to reconstruct the compressed light field. The coded color mask proposed in this work enables us to compress the color spectrum information of the light field in a single shot. Our framework achieves state-of-the-art reconstruction quality with low computational time. Moreover, our network is designed to handle multiple types of mask patterns at the same time, allowing it to decompress a light field projected at various places on the sensor and by that, avoid the usage of multiple networks and excessive memory consumption, which is the practice in previous solutions \cite{gupta2017compressive}. 

We employ a fully convolutional network (FCN) trained end-to-end using color 4D light field patches to solve the compressed sensing task. We compare our results on a light field dataset captured by a Lytro camera against dictionary-based approaches and another deep learning based method \cite{gupta2017compressive}, demonstrating the superiority of the proposed strategy. Finally, we introduce an unsupervised-trained depth estimation network, which is concatenated with our light field reconstruction network, to extract depth maps directly from the compressed measurements. With less than a second in computation time, the introduction of color compression and a memory efficient implementation, we believe that our work takes us one step closer to a real-time high-resolution light field photography. 
%---------------------------------------------------------------
\section{Related Work}
\subsection{Capturing the light field}

Light field images can be captured in various ways. For example, over a century ago they have been recorded on a film sensor using pinholes or microlens arrays \cite{ives1903parallax}. 
More recently, they have been captured using multi-device systems (e.g. array of cameras \cite{wilburn2005high}) or by time sequential imaging \cite{levoy1996light}. The problem with all these options is that they are either clumsy and expensive or unsuitable for dynamic scenes. 

To alleviate these problems other methods have been proposed. The lenslet-based approaches \cite{adelson1992single,ng2005light} use microlens array, where each of them samples the angular distribution of the light field. Light field modulation techniques \cite{veeraraghavan2007dappled,wetzstein2013plenoptic} use coded masks or apertures, with patterns such as pinholes or sum of sinusoids, to multiplex the angular information into the sensor. Yet, all these approaches sacrifice the spatial resolution of the sensor in favor of the angular resolution. 

\subsection{Compressed light field}

Compressed light field, first introduced by Marwah et al. \cite{marwah2013compressive},  allows reconstructing full resolution light field images from their coded projections. In this approach, a coded mask is placed between the sensor and the lenses to create a coded projection on the sensor. The work in \cite{marwah2013compressive} reconstructs the light field from the compressed measurements by sparse coding (using methods such as orthogonal matching pursuit (OMP) \cite{pati1993orthogonal} or the iterative shrinkage and thresholding algorithm (ISTA) \cite{beck2009fast}) 
with a learned dictionary of light field patches. Despite the fact that their solution allows high-resolution light field capturing, it suffers from huge computational complexity leading to hours of processing for a single light field scene. 

Later works suggested alternative frameworks for the capturing and/or the reconstruction tasks: Hirsch et al. \cite{hirsch2014switchable} introduced angle sensitive pixels (ASP), which create better conditions for recovery compared to mask-based approaches. They used the same time-consuming dictionary-based methods. Chen et al. \cite{chen2017light} proposed a framework in which the disparity of the scene is first calculated by sub-apertures scan, followed by reconstruction using a disparity-specific dictionary. While their framework achieved lower runtime, it is still not enough for real-time usage as it takes few minutes. Moreover, their technique is limited to static scenes. It is important to mention that none of the above frameworks dealt with color compression.   
 
\subsection{Deep learning for light field reconstruction}

Compressed sensing algorithms are usually computationally demanding, mainly because of their need for a significant amount of iterations. Gregor et al. \cite{gregor2010learning} have addressed that problem by training an encoder network, named learned ISTA (LISTA), with an architecture based on ISTA \cite{beck2009fast}. They have shown that their framework produces both better accuracy than ISTA and in much fewer iterations. Despite its benefits, LISTA is a fully connected network, which limits the patch size due to memory limitations. Moreover, a large number of training examples are required in the case of large patches. 
Therefore, an alternative neural network architecture is required.

Alternative neural network-based solutions for compressed sensing have been already suggested, e.g., for MRI reconstruction \cite{mardani2017deep} and compressed CT \cite{han2016deep}. Gupta et al. \cite{gupta2017compressive} have proposed such a method for compressed light field. They trained a two-branch network to decompress compressed light field for various sensing frameworks. One of those branches is a fully connected network, which like LISTA, limits the patch size. Vadathya1 et al. \cite{vadathya2018learning} have also designed a deep learning based approach. From the compressed image, they extract the central view and a disparity map, which are used to reconstruct the final light field by wrapping the central view. For that purpose, they use three different CNNs. Note that the network architectures in both \cite{gupta2017compressive} and \cite{vadathya2018learning} are adjusted for only one mask pattern. Therefore, they are not invariant to different locations on the sensor as each patch on the sensor is generated by a different compressing matrix.

In another work, Srinivasan et al. \cite{srinivasan2017learning} have used dilation networks in order to synthesize new viewpoints from only the central view in addition to the depth map of the scene. While they succeed to do so for specific scenes, their system relies on a large aperture camera and is limited to the class of images it was trained on.

%-------------------------------------------------------------------------

\begin{figure}[t]
\begin{center}
   \includegraphics[width=1\linewidth]{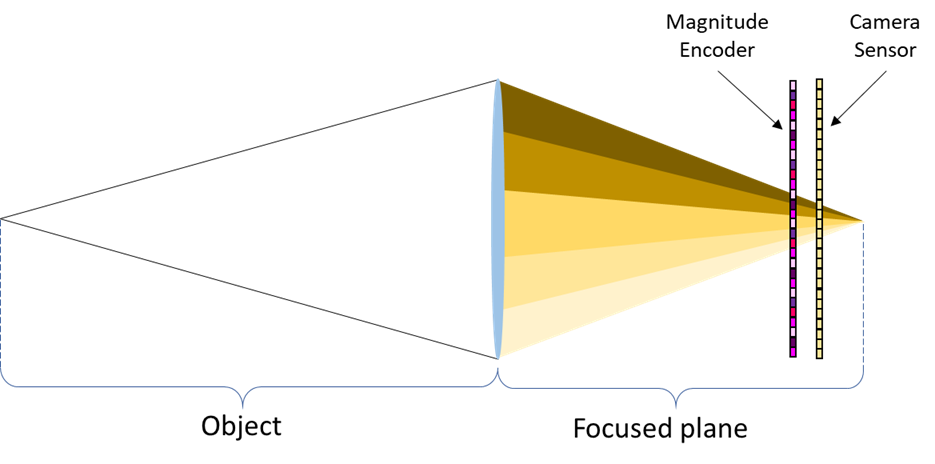}
\end{center}
   \caption{A scheme of the light field acquisition system with the color-coded random mask. Different viewpoints are coded at different places at the coded mask. Thus, each angle has a different weight in $\boldsymbol{\Phi}$ for every color channel.}
\label{fig:opticsys}
\end{figure}

\section{Color light field capturing}

In this section, we provide details on the 4D color light field images and their coded projections. 
In addition, we describe our compressed sensing problem and the properties of our proposed color-coded mask \cite{mendlovic2017system}.

Unlike the previously proposed compressed light field approaches, our system does not include CFA but only a monochrome sensor with a coded color mask located near the sensor. Therefore, the color information is also compressed in addition to the angular information but in a different way. This setup has two important advantages over using a CFA with a BW coded mask, which is used in other works: (1) It leads to a more practical implementation that uses only a single optical element instead of two. This allows implementing the compressed light field camera with only a monochrome sensor and a color mask. This advantage is crucial when building a real camera because the coded mask needs to be very close to the sensor itself and a CFA layer may prevent placing the mask in the desired location. (2) Using a single coded color mask, instead of a coded mask with a CFA, produces greater light efficiency. This is very important in view of the fact that light efficiency poses a major limitation (leads to low SNR) in the currently used compressed light field photography. Using one optical element instead of two improves this issue and makes it cheaper for production.

\subsection{Mathematical background}

Following the plenoptic multiplexing approach in \cite{wetzstein2013plenoptic} and the representation of light field in \cite{ng2005light}, we define the contiguous color light field as $l_{\lambda}(\textbf{x},\textbf{v}) = l(\textbf{x},\textbf{v},\lambda)$, which denotes the ray that intersects the aperture plane at $\textbf{x}$ and the sensor plane at $\textbf{v}$ over the color spectrum $\lambda$. 
A point at the sensor image is an integration over the aperture of all light rays that reach this point, over all the  spectrum, coded by the mask between the sensor and the aperture:
\begin{equation}
\label{eq:lf1}
i(\textbf{x})=\iint l(\textbf{x},\textbf{v},\lambda)M(\textbf{x},\textbf{v},\lambda)\cos^4\theta \,d\textbf{v}\,d\lambda, \end{equation}
where $M(\textbf{x},\textbf{v},\lambda)$ is the modulation function characterized by the coded mask and $\theta$ is the angle between the ray $(\textbf{x},\textbf{v})$ and the sensor plane. The $\cos^4\theta$ factor represents the vignetting effect \cite{ray2002applied}.
To simplify the equation, we denote: \begin{equation}
\tilde{M}(\textbf{x},\textbf{v},\lambda)=M(\textbf{x},\textbf{v},\lambda)\cos^4\theta. \end{equation}
Thus, for a specific color spectrum we get
\begin{equation}
\label{eq:lf2}
i_\lambda(\textbf{x})=\int l_\lambda(\textbf{x},\textbf{v})\tilde{M}(\textbf{x},\textbf{v},\lambda) \,d\textbf{v}.
\end{equation}

For a discrete light field, we have a vectorized version of \eqref{eq:lf2}. Taking the noise into account, we have
\begin{equation}
\label{eq:vec_lf_mono}
\textbf{i}_\lambda = \boldsymbol{\Phi}_\lambda \textbf{l}_\lambda + \textbf{n} ,\ \ \boldsymbol{\Phi}_\lambda = [\boldsymbol{\Phi}_{\lambda,1}\ \ \boldsymbol{\Phi}_{\lambda,2} \ \ \cdots \ \ \boldsymbol{\Phi}_{\lambda,N_v^2}],
\end{equation}
where $\textbf{i}_\lambda \in \mathbb{R}^m$ is the vectorized  sensor image, $\textbf{l}_\lambda \in \mathbb{R}^k$ is the vectorized light field, $\textbf{n} \in \mathbb{R}^m$ is an i.i.d zero mean Gaussian noise with variance $\sigma^2$, and $\boldsymbol{\Phi}_{\lambda,i} \in \mathbb{R}^{m\times m}$ is the modulation matrix of the ith viewpoint over the spectrum $\lambda$. $\boldsymbol{\Phi}_\lambda \in \mathbb{R}^{m\times k_\lambda}$ is a concatenation of $\boldsymbol{\Phi}_{\lambda,i}$, i.e., it is the sensing matrix based on the modulation of the projected light field at the sensor. Since $N_v$ is the angular resolution of the light field for a single axis, the discrete light field has $N_v^2$ different viewpoints. Also, if the spatial resolution of the light field is $N_x\times N_x$  then $m = N_x^2$ and $k_\lambda = N_x^2\cdot N_v^2$.

For the RGB color space $\lambda \in \{\lambda_R, \lambda_G, \lambda_B\}$ and we 
can write \eqref{eq:vec_lf_mono} as:
\begin{equation}
\label{eq:vec_lf_color}
\textbf{i} = [\boldsymbol{\Phi}_{\lambda_R} \ \boldsymbol{\Phi}_{\lambda_G}  \ \boldsymbol{\Phi}_{\lambda_B} ] \textbf{l} + \textbf{n} = \boldsymbol{\Phi}\textbf{l} + \textbf{n}, 
\end{equation}
where $\textbf{l}=[\textbf{l}_{\lambda_R} \ \textbf{l}_{\lambda_G} \ \textbf{l}_{\lambda_B}]^T $, $\boldsymbol{\Phi} \in \mathbb{R}^{m\times k}$ and $k = N_x^2\cdot N_v^2 \cdot3$.
While \eqref{eq:vec_lf_mono} is the sum of each discrete viewpoint, coded by its appropriate sensing matrix $\boldsymbol{\Phi}_{\lambda,i}$, in \eqref{eq:vec_lf_color} we have also the summation over the three color channels.  The compression ratio of the system is $\frac{m}{k} = \frac{1}{N_v^2\cdot3}$, which means that for $N_v^2 = 25$ viewpoints, the  compression ratio is $1.3\%$.
Also, the overall light which reaches the sensor is divided between each sub-aperture image among each of its color channels. Therefore, every color channel of sub-aperture image is attenuated by the same compression ratio, so the effective matrix $ \boldsymbol{\Phi} = \frac{1}{N_v^2\cdot3} \tilde{\boldsymbol{\Phi}}$, where $\tilde{\boldsymbol{\Phi}}$ is the unattenuated matrix. Due to this phenomena, the reconstruction process has higher noise sensitivity as the compression ratio increases. 

From compressed sensing perspective, the inverse problem that we wish to solve is:
\begin{equation}
\label{eq:inv_prob}
\underset{\boldsymbol{\alpha}}{\textnormal{argmin}} \ \ \| \boldsymbol{\alpha} \|_0  \ \ s.t \ \   \| \textbf{i} - \boldsymbol{\Phi}\textbf{D} \boldsymbol{\alpha} \|_2^2 < \epsilon, 
\end{equation}
where $\| \cdot \|_0$ is the $l_0$ pseudo-norm,   $\epsilon = \|\textbf{n}\|_2^2$, $\textbf{D} \in \mathbb{R}^{k \times s}$ is a given transform matrix or a learned dictionary and $\boldsymbol{\alpha} \in \mathbb{R}^s$ is the sparse representation of the light field $\textbf{l}$.  
This problem is a NP-hard problem. It can be solved using a greedy algorithm such as OMP \cite{pati1993orthogonal} or by relaxation to a $\ell_1$ minimization problem, which is also known as basis pursuit denoising (BPDN) or LASSO, and has many solvers (e.g. \cite{beck2009fast,boyd2011distributed}).% and dictionary learning  \cite{aharon2006rm,mairal2009online}.

Due to physical constraints, $\boldsymbol{\Phi}_{\lambda,i}$ are diagonal matrices and  $\boldsymbol{\Phi}$ is a concatenation of them (see Fig.~\ref{fig:phi} top). Therefore, $\boldsymbol{\Phi}\textbf{D}$ has a high mutual coherence \cite{donoho2006stable} and %does not satisfy the RIP condition \cite{Elad:2010:SRR:1895005} which means that there is 
thus, no effective theoretical guarantee for successful reconstruction. However, empirical evidence has shown that the light field images can be still restored, even without these guarantees.          

\subsection{The color mask}
\label{sec:color_mask_sec}

In order to multiplex the color information of the intersected rays into the projected 2D image at the sensor, we use a color mask, which unlike bayer CFA pattern \cite{bayer1976color}, is random. The position of the mask should also enable us to multiplex the angular information of the rays. Therefore, the mask should not be placed directly on the sensor but slightly further, as mentioned in \cite{marwah2013compressive}. This position, in addition to the color pattern, allows having random weights for different angles and colors. It is important to mention that the effective weights of $\boldsymbol{\Phi}$ cannot be taken directly from the color mask, but by an accurate ray tracing computation over all the possible $\textbf{x}, \textbf{v}$ and the three color channels. Therefore, the relationship between the mask and $\boldsymbol{\Phi}$ is not direct.

\begin{figure}[t]
\begin{center}
   \includegraphics[width=1\linewidth]{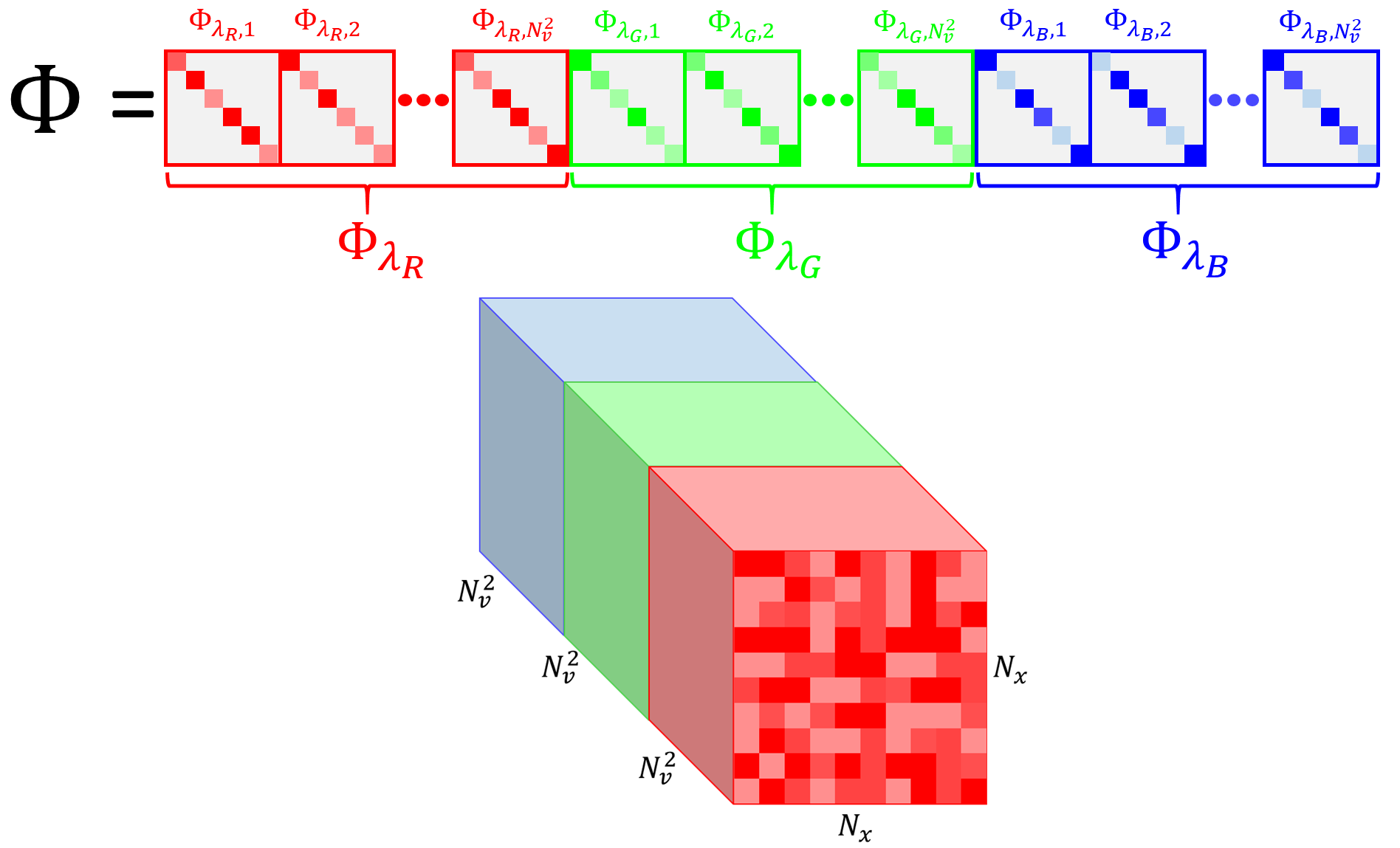}
\end{center}
   \caption{\textbf{(Top)} $\boldsymbol{\Phi}$ \textbf{matrix:} Because of physical constraints, the sensing matrix for each viewpoint and color channel is a diagonal matrix. $\boldsymbol{\Phi} \in  \mathbb{R}^{N_x^2 \times N_x^2\cdot N_v^2 \cdot3}$ is a concatenation of these matrices over all viewpoints and color channels. \textbf{(Bottom)} $\boldsymbol{\Phi}$ \textbf{tensor:} $\boldsymbol{\Phi}$ can be also expressed as a 5-D tensor. Each element in it is the modulation weight for a specific ray $(\textbf{x},\textbf{v},\lambda)$. In the above figure, we present a tensor $\boldsymbol{\Phi}$ in which the angular dimensions and the color dimension are concatenated together. This is the tensor $\boldsymbol{\Phi}$ we use as part of the input to our reconstructed network (see Fig.~\ref{fig:netarch}).} 
\label{fig:phi}
\end{figure}

A simpler approach to calculate $\boldsymbol{\Phi}$ is to illuminate the sensor from each viewpoint, using white LED and three color filters. Then, $\boldsymbol{\Phi}_{\lambda,i}$ can be easily deduced from the pattern projected at the sensor. Yet, for simplicity, in our simulation we assume $\boldsymbol{\Phi}$ to be the same as the mask. To make it as realistic as possible, we will not assume periodic mask or any specific structure of it as done in previous work \cite{gupta2017compressive,marwah2013compressive}, because as mentioned above, periodicity in the mask does not imply periodicity in $\boldsymbol{\Phi}$. Instead, we assume a random mask that implies that $\boldsymbol{\Phi}$  is also random, which is more realistic. From now on, $\boldsymbol{\Phi}$ will be a tensor of the size $N_x \times N_x \times N_v \times N_v \times 3$ (see Fig~\ref{fig:phi} bottom), where each index corresponds to the weight modulation of a specific ray $(\textbf{x},\textbf{v},\lambda$). Compression of the color and angular information is described as $\textbf{i} = \boldsymbol{\Phi}(\textbf{l}) =  S(\boldsymbol{\Phi} \odot \textbf{l})$, where $\odot$ is an element-wise multiplication and $S(\cdot)$ is a summation operator over the color and angular axes. 

We evaluated three different distributions of the color mask: (i) Uniform distribution $\boldsymbol{\Phi} \sim U[0,1]$;  (ii) RGB, where each pixel on the mask has the same probability of being red, green or blue; and (iii) RGBW, where each pixel is either red, green, blue or white (let all colors pass) with the same probability. The last two are more realistic since they are easier to be manufactured (similar to CFA). Among these distributions, RGBW produced the best results in our experiments. For more details see the supplemental material.  
%$\boldsymbol{\Phi}$ is also the compression operator of the imaging system.    

\begin{figure*}[t]
\begin{center}
   \includegraphics[width=0.8\linewidth]{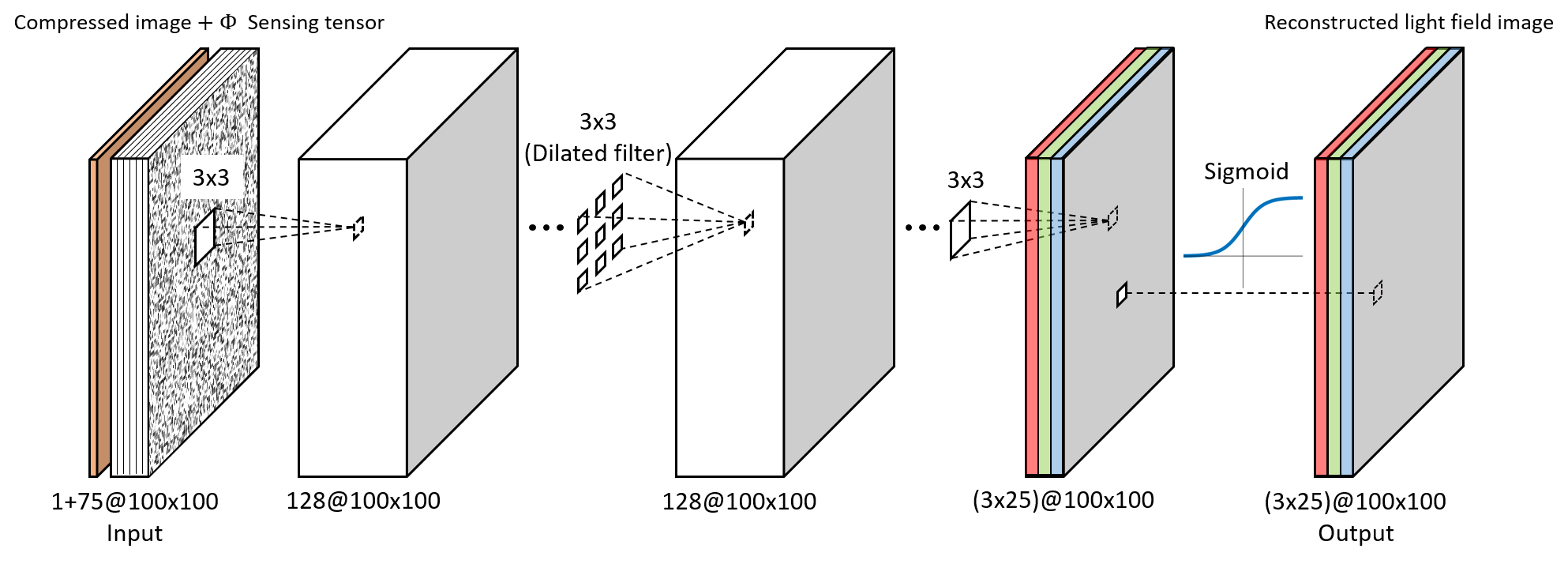}
\end{center}
   \caption{Our reconstruction network consists of $11$ $3 \times 3$ convolutional layers with 128 channels. The middle four layers are dilated convolutional layers. All the layers, except the last one, are followed by a batch normalization and ELU. The last layer is followed by a sigmoid enforcing the output range to [0, 1]. The network input is a concatenation of the compressed image and the sensing tensor (see Fig~\ref{fig:phi}). The output is the decompressed color light field, which consists of 5x5 viewpoints across 3 color channels (thus, we have 3x25 channels in the last two layers).}
\label{fig:netarch}
\end{figure*}

\section{Light Field Reconstruction Network}
In our work, we use a FCN which enables the processing of large patches (we use a size of 100x100). To make the network robust to the location of the patch in the image, it gets as an input also the corresponding part in $\boldsymbol\Phi$. This allows training a single network for the whole light field scene.
We turn to describe the network design, which allows fast reconstruction. In Section~\ref{sec:exp_sec}, we compare our network results against dictionary-based methods along with another deep learning method.

The network receives as an input the compressed image and its matching sensing tensor:
\begin{equation}
\label{eq:rec_net}
\hat{\textbf{l}} = f(\textbf{i},\boldsymbol{\Phi}),
\end{equation}
where $\hat{\textbf{l}}$ is the reconstructed light field patch, and $\textbf{i}$ and $\boldsymbol{\Phi}$ are the compressed patch and its matching sensing tensor of corresponding location at the sensor.
Due to memory limitations, in the training time our network does not process the whole images at once but in a patch based manner. During test time, we take advantage of the FCN architecture and process the whole compressed image at once.

%In order to avoid block artifacts, we reconstruct the patches with an overlap between them according to a predefined stride size. We have observed that the peripheral pixels of each patch are usually restored with less accuracy compared to the inner pixels.
%Therefore, we average the patches using Gaussian weights: 
%\begin{equation}
%\label{eq:patch_avg}
%\hat{l}(\textbf{x},\textbf{v},\lambda)=\frac{\underset{\textbf{x}_0 \in \textbf{X}}{\sum} P_{\textbf{x}_0} G_\sigma(\textbf{x}-\textbf{x}_0,\textbf{v},\lambda)}{\underset{\textbf{x}_0 \in \textbf{X}}{\sum}G_\sigma(\textbf{x}-\textbf{x}_0,\textbf{v},\lambda)},
%\end{equation}
%where $P_{\textbf{x}}$ is the recovered patch whose center is at $\textbf{x}$. $G_\sigma(\textbf{x},\textbf{v},\lambda)$ is a zero mean 2D isotropic Gaussian with variance $\sigma$ and the same support as $P_{\textbf{x}}$. %Low standard deviation of the Gaussian sets smaller weights for the outer pixels of each patch while higher standard deviation gives them more weight.

\subsection{Network architecture}
Due to their success in various tasks, we chose convolutional neural network as our regression model for the reconstruction. Convolutional networks allow us to process large patches with a low computational time. Our network architecture is a dilated convolutional network \cite{yu2015multi}. Dilated convolutions enable the expansion of the receptive fields of the network using a small filter size, without performing any pooling or sampling. This enables us to keep the original resolution of the network without making the network too deep, which may harm the computation time and lead to over-fitting. This type of network was originally created for semantic segmentation but appears to be also suitable for our compressed sensing reconstruction task as well. 

Each convolution layer in our network is followed by an exponential linear unit (ELU) \cite{clevert2015fast} and batch normalization \cite{ioffe2015batch}, except at the last layer, where a sigmoid is used in order to force the output to be in the range [0,1]. All filters are of size $3\times3$ without any exception. The dilations used in the network are with exponentially increasing rates as mentioned in \cite{yu2015multi}. We found that training with big patches leads to better reconstruction. Therefore, our network is trained with patches of size $100\times100\times5\times5\times3$ (the $5\times5\times3$ stands for the number of reconstructed angles and the color channels). 

The network has 11 convolution layers of which the middle four are dilated convolutional layers with exponentially increasing rates of 2-4-8-16. All layers have 128 channels. Our model size is 17.23 MB. As mentioned in Section~\ref{sec:color_mask_sec}, the simulated mask is randomly generated as a RGBW color mask.
 
 \begin{figure*}[t]
\begin{center}
   \includegraphics[width=0.8\linewidth]{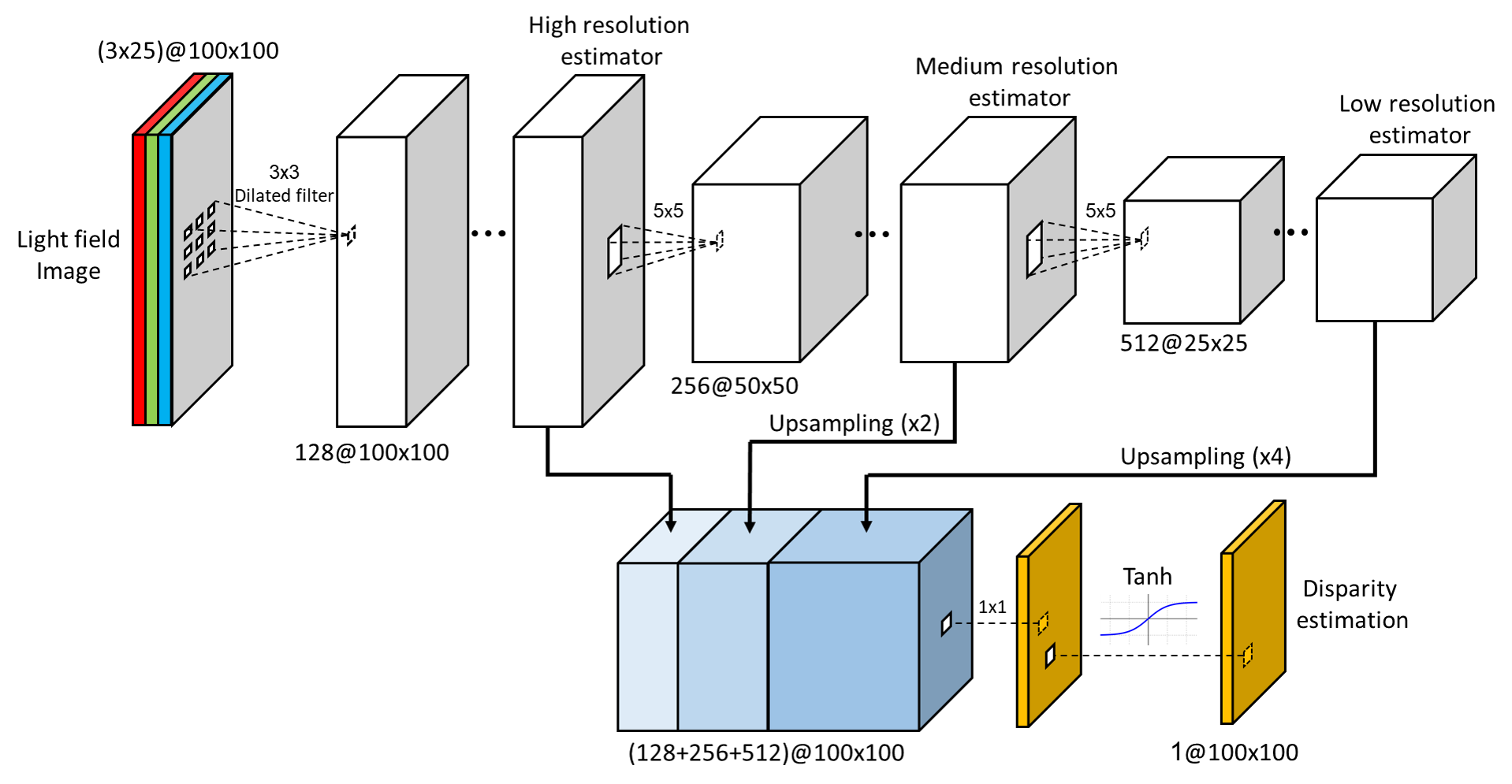}
\end{center}
   \caption{Our disparity network consists of a dilated convolutional network followed by downsampling, convolutions, another downsampling and more convolutions. Downsampling is done by a 2-stride convolution. The outputs of all resolution levels are concatenated after upsampling the medium and low-resolution maps using transposed convolutions. A $1\times1$ convolution is applied on the concatenated maps followed by a scaled tanh at the output that provides the estimated disparity field.}
\label{fig:dispnet}
\end{figure*}

\subsection{Robustness for the sensing matrix}
The input to the network is the compressed color light field patch, concatenated with its matching $\boldsymbol{\Phi}$ tensor. Adding $\boldsymbol{\Phi}$ to the input improves the reconstruction, but more importantly, allows the network to be adaptive to the distribution of the given $\boldsymbol{\Phi}$. In fact, this important property makes the network useful for reconstructing patches from different places at the sensor, which corresponds to different sensing matrices. Therefore, this allows us to train \textbf{only one} network for the whole sensor, which leads to a very small memory usage and computational time compared to the case where a different network is used for each patch.
For example, in case of 1-megapixel sensor and $100 \times 100$ patch size with $50$ pixels stride, we would need to train 324 different networks, which sums up to a size greater than 5.5GB. Our method saves all this effort and memory consumption as it allows easy manufacturing of a light field camera.
%Furthermore, we can look at the first layer's feature maps as a  learned interpolation of the compressed image with the modulation mask instead of a naive interpolation such as $\boldsymbol{\Phi}^T \textbf{i}$. 

%\begin{figure}[h]
%\begin{center}
%   \includegraphics[width=0.6\linewidth]{patchavg.png}
%\end{center}
%   \caption{Reconstructed light field patch averaging using Gaussian weights according to (10). The peripheral pixels of each patch are usually restored with less accuracy compared to the inner pixels.}
%\label{fig:long}
%\label{fig:onecol}
%\end{figure}

\subsection{Training}
We mark the set of light field patches from our training images as $\mathcal{T}$. We also create a dataset of different $\Phi$ tensors which correspond to all the locations on the sensor that we use during recovery (which are set according to a wanted stride size). We mark this data set as $\mathcal{M}$. 
For each batch of size $B$, we randomly choose light field patches from $\mathcal{T}$ and sensing tensors from $\mathcal{M}$. Then we create their matching compressed patches $\{\textbf{i}_q\}_{q=1}^{B}$:
\begin{equation}
\label{eq:train_data}
\textbf{i}_q = \boldsymbol{\Phi}_q(\textbf{l}_q) + \textbf{n}, \ \ \textbf{l}_q \in \mathcal{T}, \ \boldsymbol{\Phi}_q \in \mathcal{M}, 
\end{equation}
where the training set is the group of tuples $\{(\textbf{i}_q,\boldsymbol{\Phi}_q,\textbf{l}_q)\}_{q=1}^B$, in which every tuple consists of the ground truth light field patch $\textbf{l}_q$, its corresponding sensing tensor $\boldsymbol{\Phi}_q$ and its compressed measurement $\textbf{i}_q$. $\textbf{n}$ is the model's sensor noise $\textbf{n} \sim \mathcal{N}(0,\sigma_{sensor}^2)$.
This way we create combinations of various light field patches with randomly chosen locations in our sensing tensor.

The network training loss function consists of two parts: 
\begin{equation}
\label{eq:loss_rec}
\mathcal{L} = \underset{q}{\sum}\underbrace{\| \hat{\textbf{l}}_q-\textbf{l}_q\|_1}_{\mathcal{L}_{data}} +\beta \underbrace{\|\boldsymbol{\Phi}_q(\hat{\textbf{l}}_q)-\textbf{i}_q\|_2^2}_{\mathcal{L}_{cs}},
\end{equation}
where $\beta$ is a hyper parameter balancing $\mathcal{L}_{data}$ and $\mathcal{L}_{cs}$. The data term $\mathcal{L}_{data}$ is the $\ell_1$ distance between the reconstructed light field patch and the ground-truth patch. 
Once the network training converges, we fine tune the network using the $\ell_2$ distance in $\mathcal{L}_{data}$ instead of the $\ell_1$ norm. This action improves the recovery accuracy by 0.5 PSNR. 
We have chosen this combination of $\ell_1$ and $\ell_2$ as it has been shown to be more effective than using just $\ell_1$ or $\ell_2$ \cite{zhao2017loss}. The second term $\mathcal{L}_{cs}$ imposes consistency with the measurements model.

\begin{figure*}[t]
\begin{center}
   \includegraphics[width=1\linewidth]{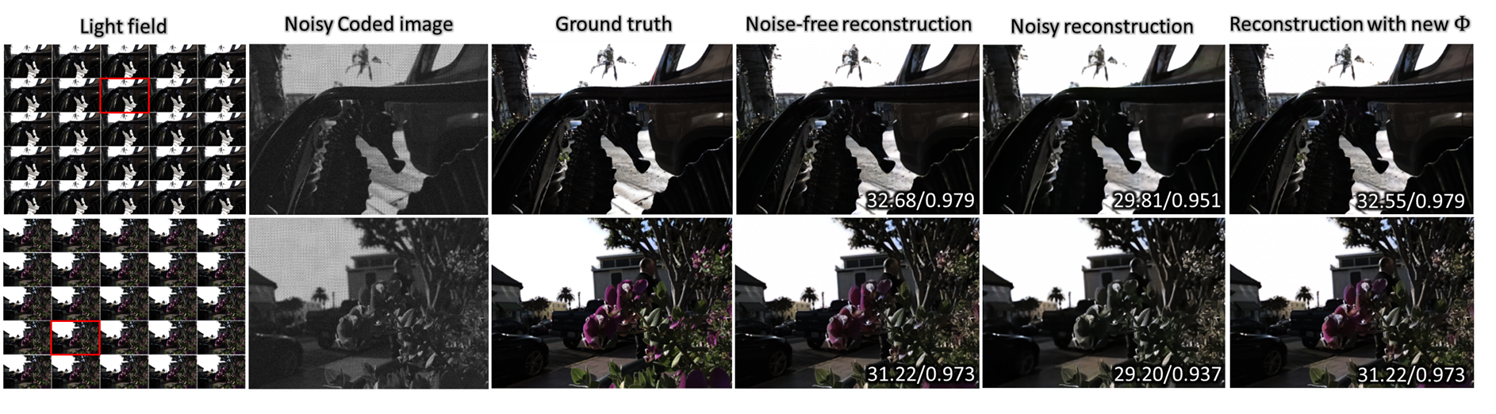}
\end{center}
   \caption{Reconstruction results on test images. Our network's reconstruction time takes less than 0.5 sec for the whole light field images. Note that our network's results are with high spatial quality and that in the noisy case they have lower color fidelity. We present also the reconstruction results for compressed light field with $\boldsymbol\Phi$ that was never been observed in training. This demonstrates the robustness of the proposed approach.}
\label{fig:sim1}
\end{figure*}

\section{Disparity Map Network}
In \cite{srinivasan2017learning}, a dilated convolutional network is proposed,  as part of the light field extraction framework, in order to extract a disparity map of the input. We use a modified version of their network and loss function. Our network reconstructs the depth map from the light field images instead of the center view as done in \cite{srinivasan2017learning}. Also, the output is a single disparity field that maps the viewpoints to the center view and not a disparity map for each angle as done in their work. 

The first 7 layers in our network consist of dilated convolutions with 128 channels and rates of 1-1-2-4-8-16-16 in order to create a prediction of the disparity map in full resolution. Then, we perform downsampling by using convolutions with stride 2, apply 3 additional convolutions with 256 channels, and then downsample again. We end by other 3 more convolutional layers with 512 channels. This way, we create two more disparity predictions with lower spatial resolution but with greater disparity reliability. 

Every layer is followed by an ELU and batch normalization. In the end, we upsample these low resolution maps with transposed convolutional layers \cite{zeiler2010deconvolutional} and concatenate them along with the full resolution map using skip connections. This action is followed by a $1\times1$ convolution layer and tanh, which is scaled to the maximum allowed disparity. All filters are 3x3 except the downsampling and upsampling layers, which are 5x5 filters and the last layer of the network, which is $1\times1$ convolution. The network's input is the light field and its output is the estimated disparity map: 
\begin{equation}
\label{eq:disp_net}
\textbf{d} = g(\textbf{l}).
\end{equation}

The loss function is similar to \cite{srinivasan2017learning} but with one crucial difference. Instead of rendering the center viewpoint to all other angels. We render all viewpoints to the center view using the disparity map at the output. This change reduce the redundancy of the disparity tensor and enforce more accurate estimation, which is based on every angle instead of just one. Due to the fact that we have only one disparity map, we do not need to use consistency regularization of the disparity map as in \cite{srinivasan2017learning}. Our loss function is:  
\begin{equation}
\mathcal{L}_D = \underset{q}{\sum} \frac{1}{N_v^2} \overset{N_v^2}{\underset{i=1}{\sum}} \| \tilde{\textbf{l}}_q^{i}-\textbf{l}_q^c\|_1+\gamma\ell_{\text{TV}}(\textbf{d}_q),
\label{eq:loss_disp}
\end{equation}
where $q$ is the patch index at the training batch $\{\textbf{l}_q\}_{q=1}^B$, $\tilde{\textbf{l}}_q^i$ is the ith viewpoint rendered to the center view, using the estimated disparity field $\textbf{d}$, and $\textbf{l}_q^c$ is the ground-truth center view. 
The rendered light field image is constructed by using the next connection  (see more details in \cite{srinivasan2017learning}):
\begin{equation}
\label{eq:depth_consis}
\tilde{l}^i(\textbf{x},\lambda)=l(\textbf{x}-\textbf{k}d(\textbf{x}),\textbf{k},\lambda).
\end{equation}
This formula is due to the depth consistency of the same point across different viewpoints without considering occlusions or non-Lambertian effects. $\textbf{k}$ is the corresponding viewpoint position of the ith index out of the total $N_v^2$ viewpoints. Notice that like \cite{srinivasan2017learning}, there is no usage of the ground truth depth map.
This network can be concatenated with the reconstruction network. Thus, a disparity map can be estimated directly from the compressed image: $\textbf{d} = g(f(\textbf{i},\boldsymbol{\Phi}))$.

\section{Experiments}
\label{sec:exp_sec}

\begin{figure*}[t]
\begin{center}
   \includegraphics[width=0.9\linewidth]{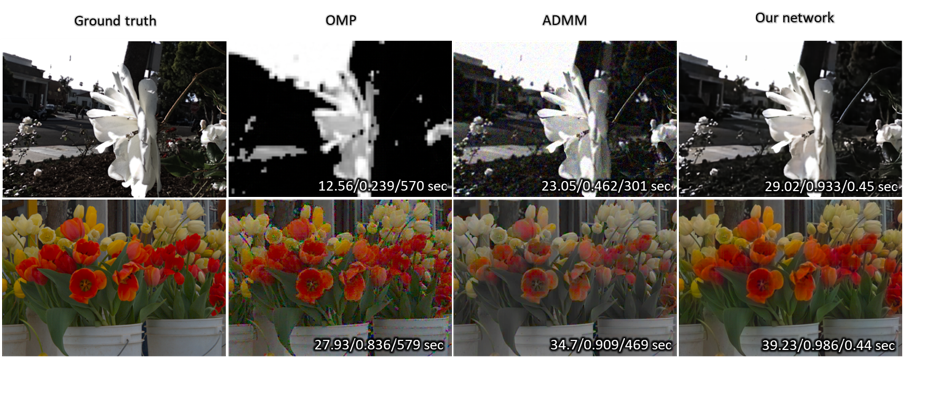}
\end{center}
   \caption{Comparison between our reconstruction network and dictionary-based methods. The presented numbers are PSNR/SSIM/reconstruction time. \textbf{Top:} noisy case with $\sigma_{sensor} =0.02$ . \textbf{Bottom:} Noiseless case. Note that our method is both faster and more accurate than the sparse coding based approaches.}
\label{fig:sim2}
\end{figure*}

We present our results for color light field decompression on the Stanford Lytro dataset \cite{stanfarch} and the Lytro dataset provided by Kalantari et al. \cite{kalantari2016learning}. In addition, we show our reconstructed disparity maps from the compressed image. All of our networks have been trained in tensorflow using color patches of size $100\times100$ with an angular resolution of $5\times5$ (25 viewpoints). The training has been done with mini-batches of size 32, the filters have been initialized using Xavier initialization \cite{glorot2010understanding}, and  the ADAM optimizer \cite{kingma2014adam} has been used with $\beta_1 = 0.9, \beta_2 = 0.999$ and an exponentially decaying learning rate. The dataset includes 79 light field images of size $376 \times 541 \times 5 \times 5 \times 3$, which means that there are over 9 millions $100\times100$ patches in it, which are different by at least a single pixel from each other. We have chosen 7 light field images as the test set while the rest have been kept for training. 

For the reconstruction network, we set $\beta = 0.004$ and the initial learning rate to $0.0005$. While for the disparity network, we set $\gamma = 0.1$ in \eqref{eq:loss_disp} and the initial learning rate to 0.001.

\subsection{Light field reconstruction evaluation}
We evaluated two scenarios. One with clean observations and the second with noisy ones with $\sigma_{sensor}=0.02$. We present the average PSNR and SSIM across our 7 test light field images. 
Both networks were trained with 800 epochs, each includes randomly chosen 4000 patches. $\boldsymbol{\Phi}$ was randomly chosen without any optimization as we describe in section \ref{sec:color_mask_sec}. The reconstructed light field patches were restored with their matching $\boldsymbol{\Phi}$ tensors using only one network for the whole image. 
We compare our results with dictionary-based methods. The dictionary was trained on color light field $8 \times 8$ patches. It was trained using online dictionary learning \cite{mairal2009online} in order to overcome memory limits, which was initialized using K-SVD \cite{aharon2006rm} trained on a smaller dataset. The reconstruction was made with OMP \cite{pati1993orthogonal} and ADMM \cite{boyd2011distributed} with patch overlapping of 2 pixels, using Intel i7-6950X CPU with 10 cores. Our network used NVIDIA GeForce GTX 1080 Ti both for training and testing.

Figure~\ref{fig:sim1} presents the quality of our reconstruction for two light field images out of our test set. Note the high accuracy in the reconstruction of the various details in the image.  
Yet, in the noisy case, we suffer from lower color fidelity because of the high compression ratio. Nevertheless, in this case, the reconstructed images have high spatial quality.
To check the ability of our network to generalize to new compression patterns of $\boldsymbol\Phi$, we have tested the network with an entirely new randomly generated $\boldsymbol\Phi$, whose patterns have never been observed by the network in training time. Our results show that switching to the new $\boldsymbol\Phi$  has not affected the results at all. This approves that our network generalizes well to new compression patterns, which are generated from the same distribution it was trained with.      

Figure~\ref{fig:sim2} makes a comparison between our network and sparsity-based methods (OMP and ADMM) for other two light field images from the test set. At the noiseless example (bottom), our reconstruction has the highest color fidelity and spatial quality, while the other methods suffer from artifacts and low saturation of the colors. In the noisy case (top), OMP fails to recover the details and colors of the image almost completely while losing all the angular resolution (shown in the supplementary material). Also, ADMM suffers from noisy reconstruction while our network's output is clean and has high reconstruction quality. On top of all that, our method takes 3 orders of magnitude less time than the sparsity methods.

Tables~\ref{tab:tab1} and Table~\ref{tab:tab2} summarize the average PSNR and SSIM both for the noisy and noiseless cases. In the noiseless case, we also compare to Gupta et al. \cite{gupta2017compressive}.
It can be clearly seen that our framework is superior in terms of reconstruction quality, computational time and has higher robustness to noise. The average reconstruction time of our network in both cases  is\textbf{ 0.15 sec} for a single light field scene, which is faster by 2-3 orders of magnitude compared to other previous existing solutions. 

According to the reported results in \cite{gupta2017compressive}, the reconstruction of noise-free images with their network on 3 reported images (Seahorse, Purple flower and White flower) has worse PSNR than ours on the same images and takes almost two order of magnitude more time compared to our reconstruction using Titan X GPU (reported results are in Table~\ref{tab:tab1}).
Note that their network does not deal with color compression but decompress each channel separately. Therefore, our compression ratio is three times lower than theirs, which is harder to decompress. Also, their network is not robust to different types of $\boldsymbol{\Phi}$  but is adjusted to only one patch pattern, which means that they have to train a different network for each patch. Note that they use ASP \cite{hirsch2014switchable}, which creates a different $\boldsymbol\Phi$ compared to coded mask.   

Since light field reconstruction is best evaluated by watching a sequence of reconstructed light field images at different angles, we have upload a video with some examples to \url{https://www.youtube.com/watch?v=F_OU1nz4V2M&feature=youtu.be} in order to demonstrate the quality of our recovery.

\begin{table}
\begin{center}
 \begin{tabular}{|c c c c|} 
 \hline 
  & \bf PSNR & \bf SSIM & \bf Reconst. time\\ [0.5ex] 
 \hline
 Ours  & 32.05 & 0.98 & 0.15 sec\\
 \hline
 OMP & 18.61 & 0.71 & 562 sec \\
 \hline
 ADMM &27.29 &0.88 & 345 sec \\
 \hline
 Gupta et al. \cite{gupta2017compressive} &30.9* & - &80 sec \\
 \hline
\end{tabular}	
\caption{Average results of 3 images reported in \cite{gupta2017compressive} for the  Lytro test set in the noiseless case. We compare our results against dictionary-based methods and \cite{gupta2017compressive}.\\
*In \cite{gupta2017compressive}, the reconstruction is made for each color channel separately and compression is performed using ASP.}
\label{tab:tab1} 
\end{center}
\end{table}

\begin{table}
\begin{center}
 \begin{tabular}{|c c c c|} 
 \hline 
  & \bf PSNR & \bf SSIM & \bf Reconst. time\\ [0.5ex] 
 \hline
 Ours & 28.9 & 0.93 & 0.15 sec  \\
 \hline
 OMP & 12.93 & 0.17 & 556 sec \\
 \hline
 ADMM & 24.16 &0.52 & 294 sec \\
 \hline
\end{tabular}	
\caption{Average results in the noisy case. We compare our results against sparsity-based methods.}
\label{tab:tab2} 
\end{center}
\end{table}

\begin{figure*}[t]
\begin{center}
   \includegraphics[width=0.9\linewidth]{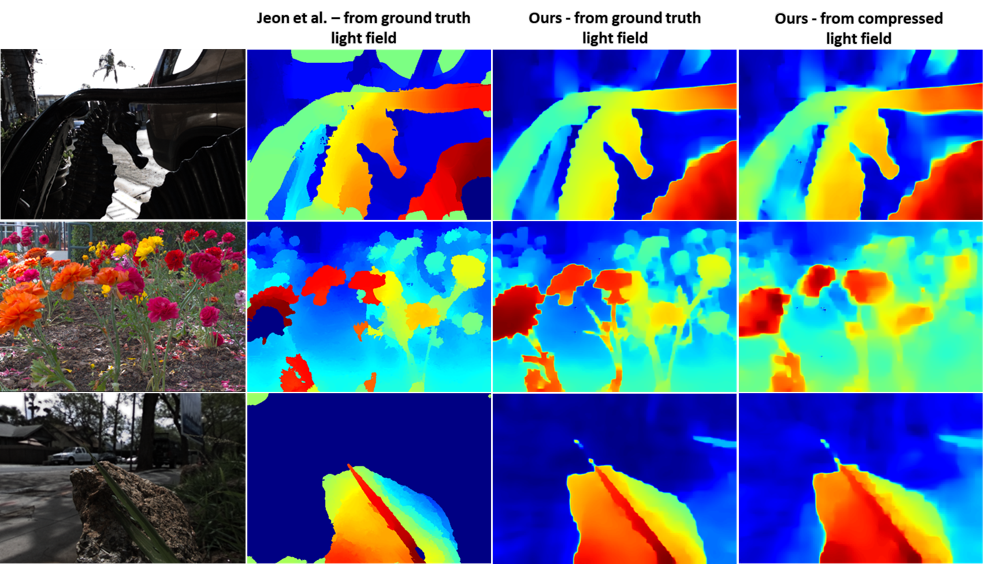}
\end{center}
   \caption{Comparison of disparity estimation. Warmer colors indicate greater disparity. We present here disparity maps that were calculated from the ground truth light field image and also from the reconstructed image. We compare our results to Jeon et al. \cite{Jeon_2015_CVPR}, which is considered to be the state-of-the-art at depth estimation from light field images. Their algorithm is demonstrated on the ground truth light field only. Note that our network produces more detailed and accurate disparity maps but with more background noise and blur compared to Jeon et al.}
\label{fig:sim3}
\end{figure*}

\subsection{Disparity estimation evaluation}
In order to evaluate our disparity network, we examine its depth estimation quality given the ground truth light field and also given the recovered light field of our reconstruction network. We compare our disparity maps to the disparity estimation of Jeon at el. \cite{Jeon_2015_CVPR}, which is considered to be the state-of-the-art for disparity evaluation from light field images. Their method uses no learning and relies on graph-cuts of a big cost volume. 

Fig.~\ref{fig:sim3} shows the recovery of each method. Our network succeeds to preserve more image details and structure. Moreover, it has less false predictions in the background pixels. Yet, it suffers from noisy background and blur in the disparity. In computation time, our network requires less than a second to calculate the disparity map using NVIDIA GeForce GTX 1080 Ti while Jeon et al. technique takes over 80 seconds using Intel i7-6950X CPU with 10 cores.

\section{Conclusions}
A novel system for reconstructing a color light field from compressed measurements coded with a random color mask has been proposed. The processing is performed by an efficient neural network that uses a small amount of memory, has low computational time and is robust to different compression patterns. We also have demonstrated how a reconstructed light field can be used to estimate the depth of the scene using another neural network. We believe that this framework can be translated into a real compressed color light field camera, which is mobile, small and cheap.

{\small
\bibliographystyle{ieee}
\bibliography{egbib}
}

%%%%%%%%%%%%%%%%%%%%%%%%%%%%%%%%%%%%%%%%%%%%%%%%%%%%APENDIX%%%%%%%%%%%%%%%%%%%%%%%%%%%%%

\onecolumn
%\renewcommand\appendixname{Appendix}
%\renewcommand\appendixpagename{Appendix}
%\setcounter{section}{0}
%\appendixpage
\appendices 
\appendix
\section{Color mask experiments}
In a real light field camera, the relationship between the mask and $\boldsymbol{\Phi}$ is not direct, which means that we cannot choose $\boldsymbol{\Phi}$ as we wish. Therefore, we limit ourself to random $\boldsymbol{\Phi}$ only. Nevertheless, we still can choose the distribution from which we generate $\boldsymbol{\Phi}$ and observe its effect on the reconstruction process. In our experiments we simulate three distributions: (i) Uniform distribution $\boldsymbol{\Phi} \sim U[0,1]$, (ii) RGB, which means every pixel on the mask has the same probability to be either red, green or blue; and (iii) RGBW, which means every pixel is either red, green, blue or white (all colors pass) with the same probability. 

For each distribution, we train our reconstruction network with the generated $\boldsymbol{\Phi}$ as mentioned in the main paper for the noise-less case. The network succeeds to adapts itself to each of the above distribution, which demonstrates the effectiveness of our method. Among these distributions, RGBW produces significantly better results (see Table~\ref{tab:tab3}). RGBW also allows for greater light efficiency, which is important in the noisy case. In addition, We test each network on a different distribution of which it was not trained on. The network, which is trained on RGBW distribution has succeeded to generalize well to the other distribution, while the RGB network couldn't generalize as well to other distributions. The network which was trained on uniform distributed $\Phi$ succeeded to generalize to RGBW data well while having more difficulty to do so for RGB. These results are expected because the RGBW and Uniform masks transmit $50\%$ of the light/information while the RGB mask transmits only third of it. %Thus, it is easier for the network to generalize between RGBW and random masks while it is easier for it to generalize from RGB to the others and vice versa. 
These results point that RGBW mask is superior not only in terms of reconstruction quality but also the ability of the network to generalize to other distribution.                   

\begin{figure}[h]
\begin{center}
   \includegraphics[width=0.8\linewidth]{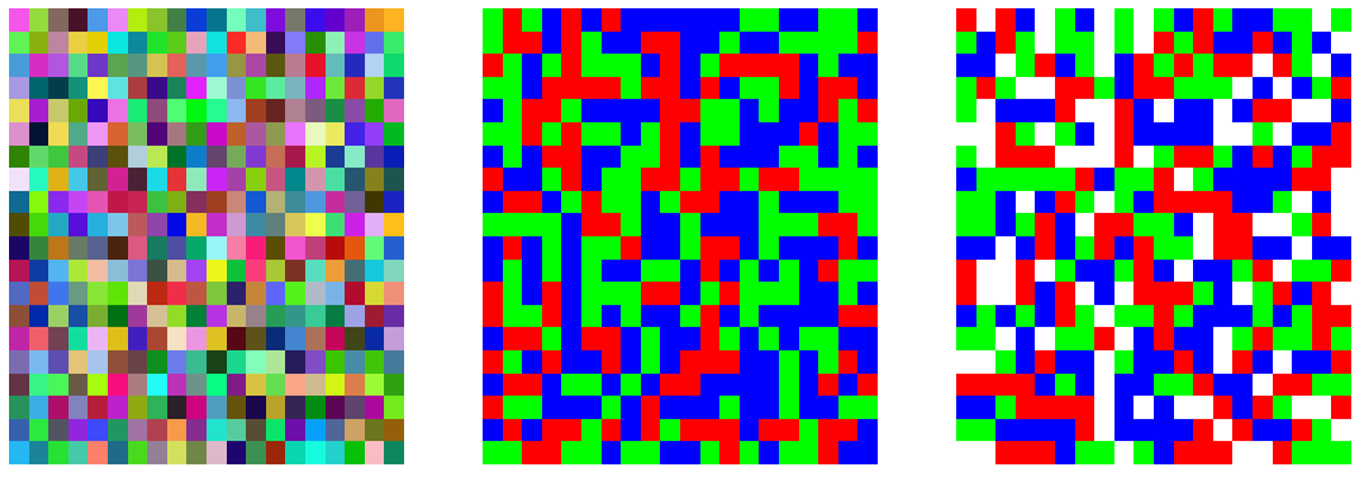}
\end{center}
   \caption{An $20 \times 20$ examples of the generated masks. \textbf{left:} uniform distribution, \textbf{center:} RGB distribution, \textbf{right:} RGBW distribution}
\label{fig:mask}
\end{figure}

\begin{table}[h]
\begin{center}
 \begin{tabular}{|c |c |c |c|} 
 \hline 
 \backslashbox{\textbf{Train data}}{\textbf{Test data}} & \bf Uniform & \bf RGB & \bf RGBW \\ [0.5ex] 
 \hline
 \bf Uniform & 29.61/0.96 & 26.97/0.90 & 29.41/0.95\\
 \hline
 \bf RGB & 23.84/0.84 & 29.79/0.94 & 22.29/0.77\\
 \hline
 \bf RGBW & 30.8/0.96 & 27.46/0.91 & \colorbox{green}{32.30/0.97} \\
 \hline
\end{tabular}    
\caption{Average results for our validation set presented as PSNR/SSIM. Each row corresponds to a different network, which was trained with a certain $\Phi$ distribution. Each column corresponds to distribution in the color mask used to create the test data. Notice that using a RGBW mask with a network trained on data from this distribution produces the best results.}
\label{tab:tab3} 
\end{center}
\end{table}

\FloatBarrier

\section{Additional light field reconstruction examples}

\begin{figure}
\begin{center}
   \includegraphics[width=1.05\linewidth]{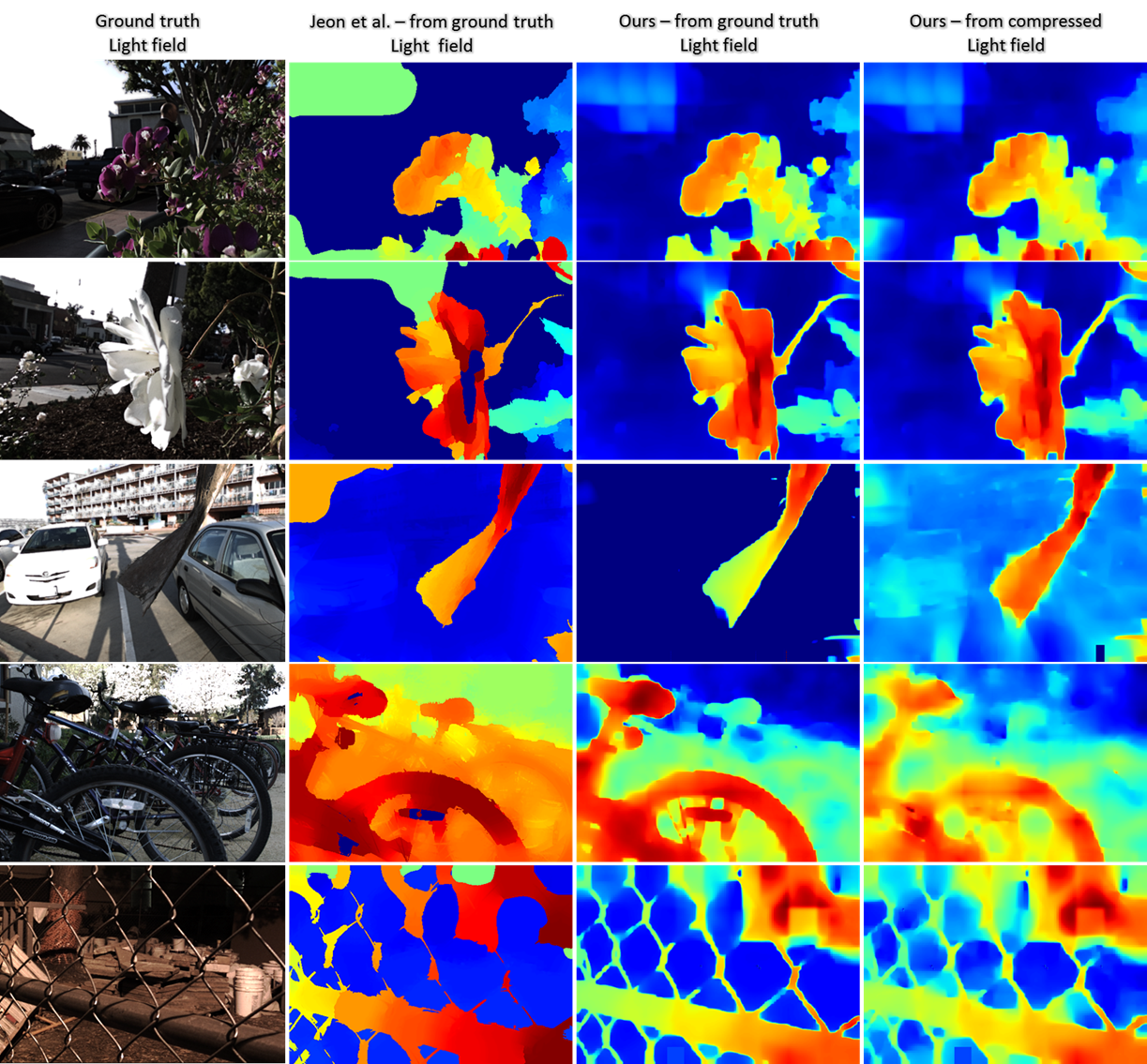}
\end{center} 
\caption{Additional depth estimation examples. We compare the results of Jeon et al. method \cite{Jeon_2015_CVPR} with our network for the ground truth light field. Also, we present our network disparity estimation from the reconstructed light field using our reconstruction network.}
\label{fig:disp}
\end{figure}

Here we present more results from our test set and other light field images from Stanford Lytro archive \cite{stanfarch}, which are not included in either the training or test set. In general, the images are taken from Stanford Lytro dataset \cite{stanfarch} and the Lytro dataset provided by Kalantari et al. \cite{kalantari2016learning}. We present results for the noiseless and noisy cases. Each example includes our full light field reconstruction and a close-up of the mark area from the four corner viewpoints of the reconstruction of each of the following methods: OMP, ADMM and ours. In each example, we marked an area with a significant disparity in the original light field. Note that besides the noise and low-quality resolution in the sparsity-based reconstruction, there is also a loss of the angular resolution. It can be well observed in the OMP recovery of the White flower and Seahorse images in the noisy case (Figs.~\ref{fig:5} and~\ref{fig:6}). OMP recovery loses the difference between the angles. Notice that in the noisy case, all techniques including ours encounter hardships in the color restoration.  Due to high compression ratio, the noise causes errors in the color fidelity of the reconstructed light field (see Fig.~\ref{fig:7}). Nevertheless, our reconstruction still has high spatial and angular resolution in this case, which can be used for depth estimation (see Fig.~\ref{fig:disp}), which is quite good. Also, it still outperforms the sparsity-based methods, which provide poor recovery. 

\begin{figure}[h]
\begin{center}
   \includegraphics[width=1.05\linewidth]{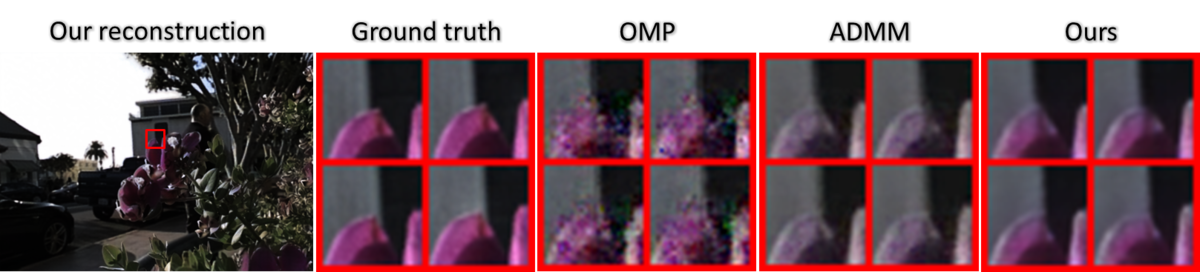}
\end{center}
   \caption{Purple flower}
\label{fig:1}
\end{figure}

We also present the three images that we used for comparing to Gupta et al. \cite{gupta2017compressive} in the main paper: Seahorse, White flower and Purple flower (Figs.~\ref{fig:1},~\ref{fig:5} and~\ref{fig:6}). Since Gupta et al. did not provide any code, we cannot present their results but according to their reports, the average PSNR of the three images is 30.9 while ours is 31.56 in the noiseless case. Note that their network does not deal with color compression but decompress each channel separately. Therefore, our compression ratio is three times lower than theirs. They also use ASP \cite{hirsch2014switchable}, which creates a different $\boldsymbol\Phi$ compared to the coded mask approach we use. 

\subsection{Noiseless case examples}

\begin{figure}[h]

\begin{center}
   \includegraphics[width=1.05\linewidth]{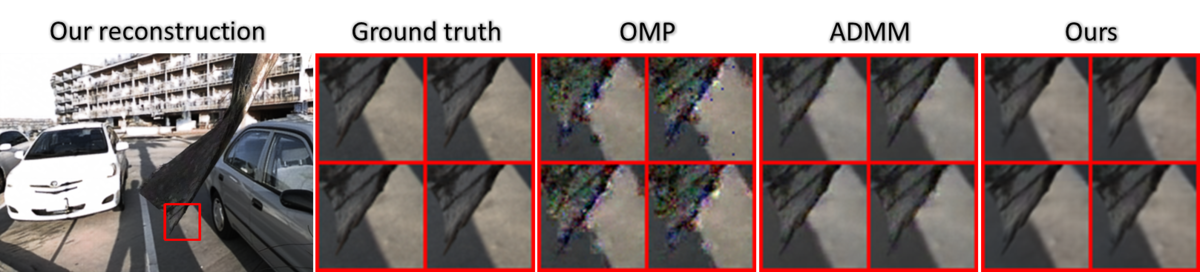}
\end{center}
   \caption{Cars}
\label{fig:2}
\end{figure}

\begin{figure}[h]
\begin{center}
   \includegraphics[width=1.05\linewidth]{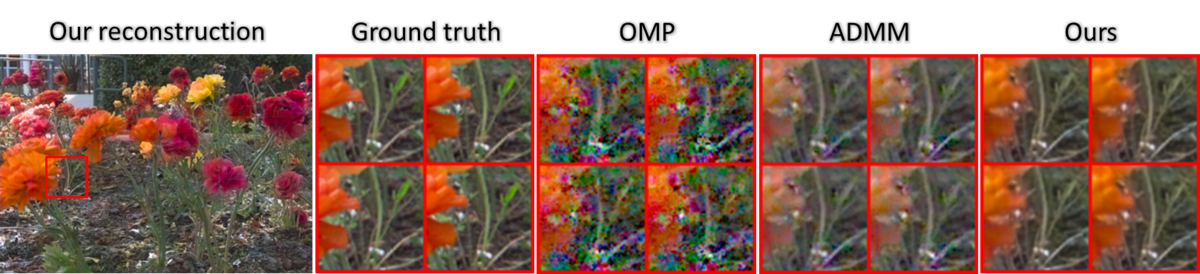}
\end{center}
   \caption{Garden}
\label{fig:3}
\end{figure}

\begin{figure}[h]
\begin{center}
   \includegraphics[width=1.05\linewidth]{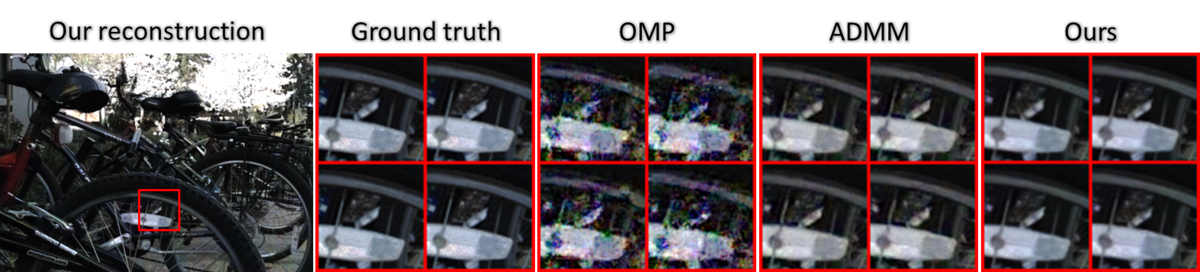}
\end{center}
   \caption{Bicycles}
\label{fig:4}
\end{figure}

\FloatBarrier
. \\ \\ \\ \\ \\ \\ \\ \\ \\ \\
\FloatBarrier
\subsection{Noisy case examples}
\begin{center}
\begin{figure}[h]
\begin{center}
   \includegraphics[width=1.05\linewidth]{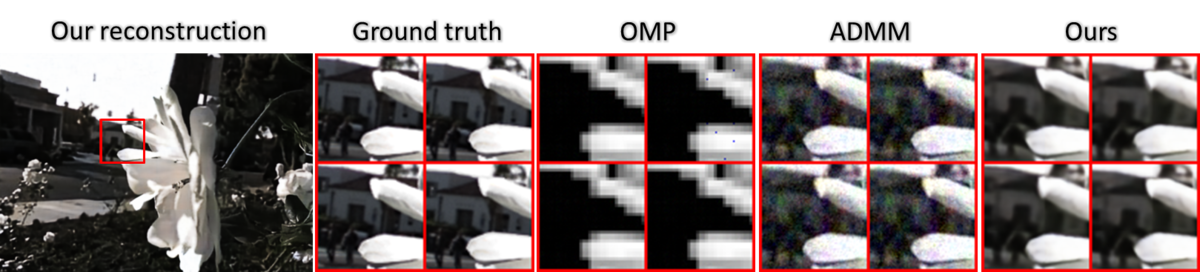}
\end{center}
   \caption{White flower}
\label{fig:5}
\end{figure}

\begin{figure}[h]
\begin{center}
   \includegraphics[width=1.05\linewidth]{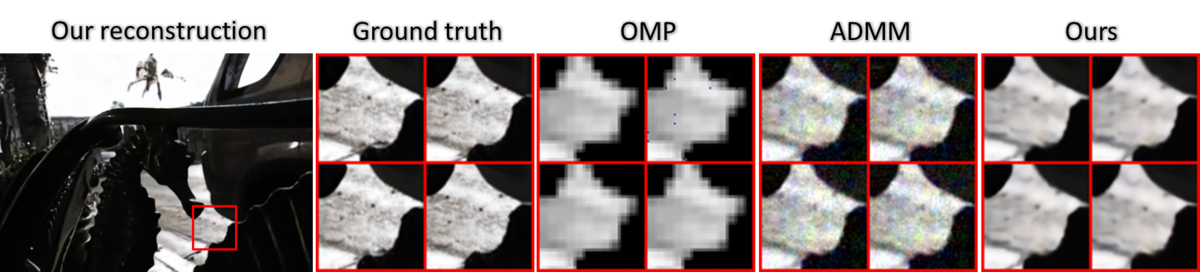}
\end{center}
   \caption{Seahorse}
\label{fig:6}
\end{figure}

\begin{figure}[h]
\begin{center}
   \includegraphics[width=1.05\linewidth]{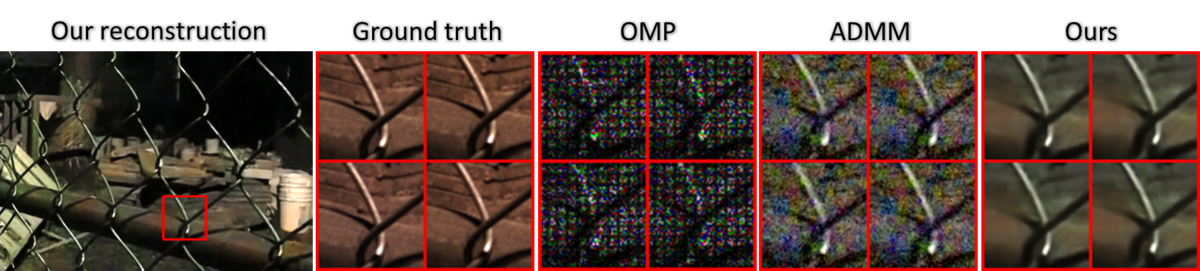}
\end{center}
   \caption{Fence}
\label{fig:7}
\end{figure}
\end{center}
\FloatBarrier

\section{Additional depth estimation examples}
We present additional disparity map estimations images from the last section that do not appear in the main paper. For every light field scene we present (i) the ground truth image; (ii) Jeon at el. \cite{Jeon_2015_CVPR}; (iii) our network disparity map from the ground truth light field; and (iv) our network disparity map from the estimated light field calculated by our light field reconstruction network (noiseless case). Note that our network provides more accurate disparity maps compared to \cite{Jeon_2015_CVPR} when using the ground truth light field. When using the reconstructed one, the disparity maps from the compressed light field suffer from some blur and artifacts, both in spatial and angular dimensions due to estimation errors at the reconstructed light field, (as can be seen in Figs~\ref{fig:1}-\ref{fig:7}). Yet, this disparity estimation is quite accurate and competitive with the one achieved by \cite{Jeon_2015_CVPR}.   

%---------------------------------------------------------------

\end{document}